\documentclass[11pt]{article}
\usepackage[margin=1in]{geometry}
\usepackage{amsmath,amssymb,amsfonts}
\usepackage{booktabs}
\usepackage{graphicx}
\usepackage{natbib}
\usepackage{url}
\usepackage{microtype}
\usepackage{xcolor}
\usepackage{caption}
\usepackage{tikz}
\usetikzlibrary{arrows.meta,positioning,shapes.geometric}
\usepackage[hidelinks,breaklinks=true]{hyperref}
\hypersetup{
  pdftitle={Pitwall: Faithful Natural-Language Race-Strategy Briefings from a Calibrated Real-Time Monte Carlo Engine},
  pdfauthor={Juan S. Santillana},
  pdfsubject={Faithful, verifier-gated natural-language generation grounded in a calibrated Formula 1 strategy simulator},
}

\newcommand{\pass}{\mathrm{pass}}
\newcommand{\ind}[1]{\mathbf{1}\!\left[#1\right]}

\title{\textbf{Pitwall: Faithful Natural-Language Race-Strategy Briefings\\
from a Calibrated Real-Time Monte Carlo Engine}}

\author{%
Juan S.\ Santillana\\
Independent Researcher\\
\texttt{juan@jsantillana.com}
}
\date{}

\begin{document}
\maketitle

% -------------------------------------------------------------------------
\begin{abstract}
Live sports commentary is grounded generation under a deadline: statements
concern real, named athletes, the grounding state changes every few seconds,
and no reference text exists at generation time. We present
\textsc{Pitwall}, a production system that generates natural-language
Formula~1 strategy briefings in English, Spanish, and Portuguese, and that
treats faithfulness as an \emph{architectural} property rather than an
aspiration: every published sentence is decomposed into typed factual
claims (positions, gaps, tyres, pace, overtakes, race control) and each
claim is verified against the probabilistic race state that prompted it.
The same verifier gates the fine-tuning data---of 3{,}045
model-written targets, only the 81.9\% whose every claim is
state-supported are retained; the rest fall back to a provably faithful
template---so the generator is trained never to have seen an ungrounded
target. What makes verification \emph{meaningful} is the grounding
substrate: a vectorized Monte Carlo engine ($N{=}2{,}000$ per-lap race
continuations) whose probabilities are calibrated on 126 races
(2018--2024) and validated on fully held-out 2025--2026 seasons
(winner-in-top-3 $90.3\%$ over 155 backtests; held-out Brier $0.0745$).
A recurring finding spans both halves of the system: virtues trade off and
must be gated separately. In simulation, calibration-optimal is not
decision-optimal; in generation, fine-tuning on richer targets buys
vividness that collapses into hallucination precisely when the grounding
state is sparse---a collapse that a four-base replication traces to the
base model's instruction adherence rather than to scale, and that
base-model selection under a sparse-context audit removes in
production. End-to-end operation---live timing to verified
trilingual briefings---was confirmed at two consecutive live Grands Prix
(Austria and Britain, 2026); at Silverstone a timestamped probability
trace, committed to disk before the outcome was known, locked onto the
eventual winner ten laps before the flag.
\end{abstract}

% -------------------------------------------------------------------------
\section{Introduction}
\label{sec:intro}

A modern Formula~1 race is decided as much on the pit wall as on the track.
A pit stop costs roughly 20--25\,s of race time, tyre compounds degrade
non-linearly and at circuit-specific rates, overtaking difficulty varies by
an order of magnitude between venues, and Safety-Car interruptions arrive
stochastically and reprice every open strategic option within a single lap.
The teams' strategists must therefore answer questions of the form
\emph{``if we box now, what is the probability we emerge ahead of the car we
are racing---and how does that change if we wait two laps?''} in well under a
minute, continuously, for two hours. Every top team operates simulation
tooling for this purpose, but that tooling is proprietary, its calibration
has never been publicly audited, and the academic literature on race
strategy simulation \citep{bekker2009planning,heilmeier2020virtual} stops
short of the real-time, probability-calibrated, decision-focused setting in
which the problem is actually solved.

The answers, moreover, are consumed as \emph{language}. Engineers,
broadcasters, and audiences do not read probability vectors; they read
sentences---``Norris pits and rejoins P4, 2.1\,s behind Sainz on
mediums''---and a generated sentence about a real, named athlete that is
confidently wrong is worse than no sentence at all. Live commentary is
therefore a demanding instance of grounded generation
\citep{wiseman2017challenges,ji2023survey}: the grounding state mutates
every few seconds, no reference text exists at generation time (ruling out
reference-based evaluation entirely), and the sparse moments---early laps,
partial state after an ingestion hiccup---are exactly the moments a fluent
generator is most tempted to fill from its prior.

This paper describes \textsc{Pitwall}, to our knowledge the first publicly
documented end-to-end system that (i) ingests the official F1 live-timing
stream during a session, (ii) maintains a canonical probabilistic race state,
(iii) evaluates strategic counterfactuals by vectorized Monte Carlo
simulation at interactive latency, (iv) reports \emph{calibrated}
probabilities---win, podium, points, and undercut success---validated against
seven complete scored seasons of historical racing (2018--2025, excl.\ 2022)
plus the opening races of 2026, and (v) verbalises state and
recommendations as natural-language briefings in English, Spanish, and
Portuguese, publishing only text whose factual claims pass verification
against the engine state (Section~\ref{sec:language}).
The system runs in production on commodity cloud infrastructure; operational
feasibility was confirmed at the live 2026 Austrian and British Grands
Prix.

Beyond the system itself, we make a methodological argument that we believe
generalizes to decision-support systems in other sports. Simulation
components are usually adopted because they make the simulator more
\emph{realistic}. We instead gate every component with two independent
held-out tests: does it improve \emph{outcome calibration} (Brier score and
reliability of finish-position probabilities), and does it improve
\emph{decision fidelity} (does the recommended strategy match what actually
worked)? These two criteria repeatedly disagree in our experiments. A
learned overtake kernel that beats a threshold baseline out-of-sample at
predicting individual passes does \emph{not} improve field-level calibration
once the simulator is re-calibrated around it; a compound-pace normalization
that demonstrably fixes pathological live pit calls \emph{worsens} finishing
order prediction, because the raw, confounded per-race fit carries genuine
race-specific signal. Rather than resolving this tension by fiat, the engine
routes each component to the path it improves: the \emph{oracle path} that
produces calibrated probabilities keeps the Brier-optimal configuration,
while the \emph{decision path} that produces recommendations uses the
feasibility-aware, normalized models. The language layer supplies a third
instance of the same lesson: fine-tuning the commentary generator on
richer, more natural targets improves fluency and coverage on
information-dense states, yet the identical model fabricates drivers,
gaps, and tyre compounds when the grounding state is sparse
(Section~\ref{sec:nlgeval}); replicating the recipe across four base
models shows this failure is a property of the base model's instruction
adherence, not of scale or of the training data---richness and
precision, like calibration and decision quality, are separate
objectives that must be gated separately, and the base model is itself
one of the gates.
We report both the positive and the
negative gating results, since the negative results---rare in applied sports
analytics---are precisely what makes the faithfulness and calibration
claims credible.

Our contributions are:
\begin{enumerate}
\item \textbf{A validated, calibrated real-time Monte Carlo engine.} A
  vectorized simulator (all $N$ races advanced simultaneously as
  $N \times C$ array operations over cars $C$) with per-lap Safety-Car and
  Virtual-Safety-Car regimes, dirty-air interaction, stochastic
  retirements, and a game-theoretic rival policy; calibrated by
  bound-constrained differential evolution on 126 training races (2018--2024)
  and evaluated with Brier score, expected calibration error, and reliability
  analysis on the fully held-out 2025--2026 seasons
  (Section~\ref{sec:calibration}, Section~\ref{sec:experiments}).
\item \textbf{A learned overtake kernel with an honest ablation.} A logistic
  model of pass probability fitted on 131{,}000 adjacent-car battles
  (4{,}142 clean passes) with per-circuit difficulty offsets, which improves
  pass prediction out-of-sample (Brier $0.0275$ vs.\ $0.0298$ baseline,
  passes-per-race MAE $8.9$) yet yields no field-calibration gain when wired
  into the simulator---an instructive negative result we analyze in
  Section~\ref{sec:h7}.
\item \textbf{The undercut window as a CRN counterfactual.} A formal
  treatment of the box-now/box-later decision as a common-random-numbers
  comparison across pit horizons $h \in \{0,1,2\}$ and a stay-out policy,
  so that inter-option probability deltas reflect strategy rather than
  sampling noise, with percentile-bootstrap confidence intervals on every
  reported probability (Section~\ref{sec:h14}).
\item \textbf{Compound-base normalization and the calibration/decision
  split.} A principled projection of fitted per-compound base paces onto the
  canonical soft$\le$medium$\le$hard ordering that eliminates a
  compound$\times$race-phase confound responsible for pathological
  ``box to the fast tyre'' live calls; together with the finding that this
  correction must live on the decision path only
  (Sections~\ref{sec:h9},~\ref{sec:dualpath}).
\item \textbf{A verifier-gated language layer with faithfulness-gated
  fine-tuning.} A typed claim schema (ten claim types over positions,
  gaps, tyres, pace, overtakes, and race control) with extraction and
  state-verification in three languages; a fine-tuning corpus of 3{,}045
  trilingual briefing targets in which a model-written target is admitted
  only if every claim it makes is supported by the race state (81.9\%
  admitted; the remainder fall back to a provably faithful template); and
  a held-out audit across four base models (1B--7B) showing that the
  resulting richness--precision trade-off is real but base-dependent:
  three bases hallucinate precisely on sparse states while a fourth
  abstains, isolating base-model instruction adherence---not parameter
  count and not the training data---as the deciding factor
  (Sections~\ref{sec:language},~\ref{sec:nlgeval}).
\item \textbf{An end-to-end live deployment.} Live-timing ingestion
  $\rightarrow$ state reconstruction $\rightarrow$ calibrated Monte Carlo
  $\rightarrow$ recommendation $\rightarrow$ verifier-grounded natural
  language commentary, demonstrated at the 2026 Austrian and British
  Grands Prix, the latter with a live probability trace scored against
  the official result (Sections~\ref{sec:live}--\ref{sec:live2}).
\end{enumerate}

% -------------------------------------------------------------------------
\section{Related Work}
\label{sec:related}

\paragraph{Race strategy simulation.}
Early work modeled circuit racing strategy as deterministic lap-time
optimization: \citet{bekker2009planning} built a discrete-event simulation
of the Zandvoort circuit for race planning, and \citet{limebeer2014optimal}
treated fuel/tyre scheduling as an optimal control problem. The most
complete academic treatment is the race simulation of
\citet{heilmeier2020virtual}, whose FAST algorithm couples lap-time,
tyre-degradation, and fuel models and was later extended with a
virtual strategy engineer based on neural networks
\citep{heilmeier2020vse}. These simulators are calibrated to reproduce
race \emph{times}; none report probability calibration of race
\emph{outcomes}, none model the box-now-vs-later counterfactual under
common random numbers, and none operate against a live timing feed.
Monte Carlo race simulation is also used commercially (e.g.\ by the teams
themselves and by broadcast providers), but no methodology or validation
is public.

\paragraph{Monte Carlo methods and forecasting in sports.}
Monte Carlo season and match simulators are standard in basketball,
football, and soccer analytics; forecast quality is typically assessed with
the Brier score \citep{brier1950verification} and its
reliability--resolution--uncertainty decomposition \citep{murphy1973new}.
Our use of percentile-bootstrap confidence intervals on simulated event
probabilities follows standard resampling practice \citep{efron1994introduction},
and our variance-reduction strategy for comparing policies under a shared
random world is the classical common-random-numbers technique from the
stochastic simulation literature \citep{glasserman1992some,law2015simulation}.
Calibration of probabilistic classifiers and forecasters more broadly is
surveyed by \citet{guo2017calibration} and \citet{gneiting2007probabilistic};
we adopt reliability diagrams and expected calibration error (ECE) from that
literature.

\paragraph{Overtaking and interaction effects in F1.}
Overtaking in Formula~1 is strongly circuit-dependent and pace-gated.
Statistical analyses of historical passing
\citep{degroote2021overtaking} document order-of-magnitude differences in
pass counts between street and power circuits, consistent with the
per-circuit offsets our kernel learns (Monaco and Imola hardest; Las Vegas,
Spa, and Qatar easiest). Aerodynamic wake (``dirty air'') penalties for a
following car are documented in the vehicle-dynamics literature and
motivated the 2022 technical regulations \citep{guerrero2020wake}; we model
them as a calibrated per-lap loss inside a gap window rather than
aerodynamically. Tyre degradation modeling, including the non-linear
``cliff,'' appears in engineering treatments of race tyres
\citep{kelly2012tyre,heilmeier2020virtual}; our contribution is mining
circuit-and-compound-specific cliff knots from seven scored seasons of stint data
with an explicit evidence gate, rather than assuming a functional form
globally.

\paragraph{Faithfulness in data-grounded generation.}
Data-to-text generation has long been evaluated by decomposing outputs
into atomic facts checked against the source records, from the extractive
metrics of \citet{wiseman2017challenges} on basketball box scores to
gold-standard fact-level protocols \citep{thomson2020gold}; in abstractive
summarization, \citet{maynez2020faithfulness} established that fluent
neural generators hallucinate content unsupported by their input, and
\citet{ji2023survey} survey the resulting hallucination literature. Our
setting sharpens these concerns in three ways: claims concern live, named
athletes; the grounding source is a probabilistic state that changes
lap-to-lap, so verification must be \emph{reference-free}; and generation
happens under a latency budget that precludes post-hoc human review. The
claim-extraction-and-verification methodology we build on, including the
finding that low-coverage extractors make faithfulness scores degenerate,
is developed in a companion paper \citep{santillana2026precision}; here we
extend it from post-race ground truth to the live state, and---to our
knowledge novel---use the verifier not only for evaluation but as an
\emph{admission gate on the fine-tuning data itself}
(Section~\ref{sec:sftgate}).

\paragraph{Decision support and human-in-the-loop deployment.}
Our deployment philosophy---calibrated probabilities with uncertainty
intervals presented to a human decision maker, never an autonomous
action---follows the forecasting-as-decision-support tradition
\citep{gneiting2007probabilistic}.

% -------------------------------------------------------------------------
\section{System Overview}
\label{sec:system}

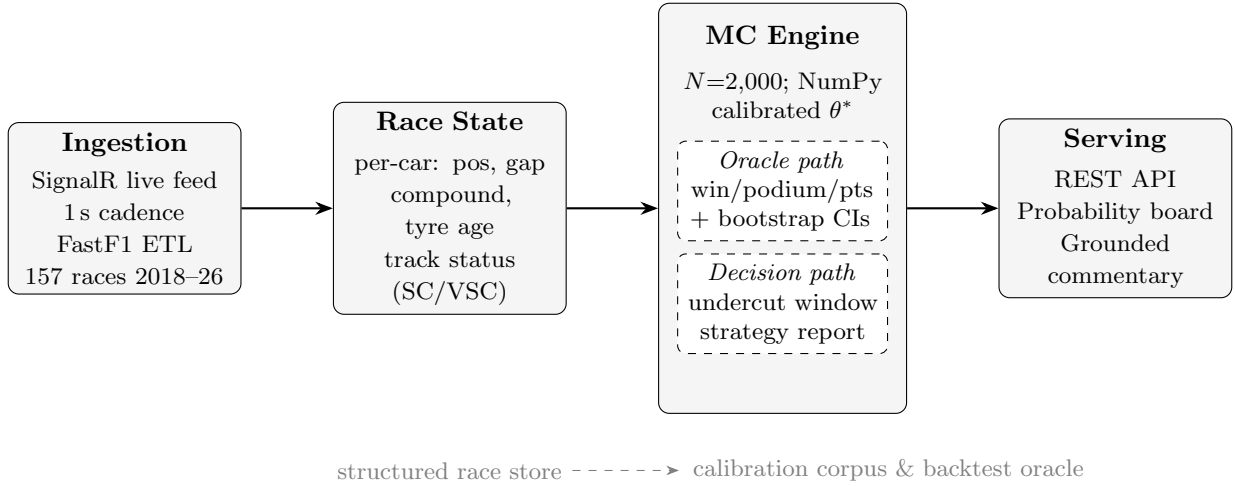
\begin{figure}[t]
\centering
\resizebox{0.98\linewidth}{!}{%
\begin{tikzpicture}[
  >=Stealth,
  box/.style={draw, rounded corners=4pt, minimum width=3.0cm, minimum height=1.8cm,
              text width=2.8cm, align=center, fill=gray!8, font=\small},
  sub/.style={draw, dashed, rounded corners=3pt, minimum width=2.6cm,
              text width=2.5cm, align=center, fill=white, font=\footnotesize},
  arr/.style={->, thick},
  node distance=1.0cm and 1.2cm
]
  % Main pipeline boxes
  \node[box] (A) {%
    \textbf{Ingestion}\\[3pt]
    {\footnotesize SignalR live feed\\1\,s cadence}\\[1pt]
    {\footnotesize FastF1 ETL\\157 races 2018--26}};
  \node[box, right=of A] (B) {%
    \textbf{Race State}\\[3pt]
    {\footnotesize per-car: pos, gap}\\
    {\footnotesize compound, tyre age}\\
    {\footnotesize track status (SC/VSC)}};
  \node[box, right=of B, minimum height=5.4cm, minimum width=3.2cm, text width=3.0cm] (C) {};
  \node[font=\small\bfseries, anchor=north] (ctitle)
    at ([yshift=-0.16cm]C.north) {MC Engine};
  \node[font=\footnotesize, align=center, text width=2.6cm, anchor=north] (cmeta)
    at ([yshift=-0.04cm]ctitle.south) {$N{=}2{,}000$; NumPy\\calibrated $\theta^*$};
  \node[box, right=of C] (D) {%
    \textbf{Serving}\\[3pt]
    {\footnotesize REST API}\\
    {\footnotesize Probability board}\\
    {\footnotesize Grounded commentary}};

  % Sub-path labels inside engine box (anchored top-down so nothing
  % covers the title or overflows the parent box)
  \node[sub, anchor=north] (oracle) at ([yshift=-0.16cm]cmeta.south) {%
    \textit{Oracle path}\\win\slash podium\slash pts\\+ bootstrap CIs};
  \node[sub, anchor=north] (decision) at ([yshift=-0.16cm]oracle.south) {%
    \textit{Decision path}\\undercut window\\strategy report};

  % Arrows
  \draw[arr] (A) -- (B);
  \draw[arr] (B) -- (C);
  \draw[arr] (C) -- (D);

  % Calibration corpus annotation below (y clear of C's tall box)
  \node[font=\footnotesize, text=gray, anchor=north] (store)
    at ([yshift=-0.5cm]C.south -| B) {structured race store};
  \draw[->, gray, dashed, thin] (store.east) -- ++(1.4,0);
  \node[font=\footnotesize, anchor=west, text=gray]
    at ([xshift=1.45cm]store.east) {calibration corpus \& backtest oracle};
\end{tikzpicture}}
\caption{\textsc{Pitwall} architecture. The same structured race store (157 races, 2018--2026,
excl.\ 2022) serves as calibration corpus, held-out backtest oracle, and grounding source for the
language layer. The MC Engine runs two parallel paths: an oracle path (Brier-optimal configuration)
for calibrated probabilities and a decision path (H5+H9 enabled) for recommendations.}
\label{fig:arch}
\end{figure}

\textsc{Pitwall} consists of four stages (Figure~\ref{fig:arch}).

\textbf{Data pipeline.} Historical data for 157 races (2018--2026; 2022 is
absent because its timing data does not load reliably) is
extracted with FastF1 and normalized into a structured JSON store: per-race
stints (compound, start/end lap, tyre life), lap times, pit stops, Safety
Car and VSC periods, and final classification. Degradation models are fit
per race and compound on \emph{green laps only}---laps under SC/VSC, in-laps,
and out-laps are excluded---yielding for each compound $t$ a base pace
$b_t$ and a linear wear slope $\delta_t$, with a fuel-burn correction
(Section~\ref{sec:laptime}). During a live session the same state is
reconstructed incrementally from the official SignalR live-timing stream,
received through a low-latency proxied edge agent and decoded into the
identical canonical schema, so that the live engine and the backtest engine
run the same engine code path.

\textbf{Engine.} The simulator advances all $N$ Monte Carlo replications of
the remaining race in lockstep as array operations (Section~\ref{sec:mc}).
Model components beyond the base engine are individually flag-gated
(Table~\ref{tab:components}); each flag was adopted, shelved, or routed to
the decision path based on held-out evidence (Section~\ref{sec:dualpath}).

\textbf{Decision layer.} On top of raw finish distributions the engine
exposes: win/podium/points probabilities with 90\% percentile-bootstrap
confidence intervals; the H14 undercut window (box now vs.\ $+1$ vs.\ $+2$
laps vs.\ stay out, under common random numbers); undercut/overcut verdicts
against a named rival modeled as a pace-optimal open-loop agent (H1:
executes its pre-race strategy at the current pace estimate, non-reactive to
the ego car's stops); and a strategy-cost report comparing each driver's
executed plan to the feasibility-constrained optimum.

\textbf{Serving.} A REST API on Azure Container Apps serves the live
probability board and decision cards; a fine-tuned language model verbalises
selected engine outputs as grounded commentary, with each generated claim
checked against the engine state by the verifier before publication.

\begin{table}[t]
\centering\small
\begin{tabular}{@{}llll@{}}
\toprule
ID & Component & Fitted on & Path (evidence) \\
\midrule
H1  & Pace-optimal rival policy        & --- (assumption)          & oracle + decision \\
H5  & Tyre cliff + feasibility caps    & 8{,}278 stints / 191k laps & decision (stops-match $+9.5$pp) \\
H7  & Logistic overtake kernel         & 131k battles / 4{,}142 passes & shelved for oracle; live narrative \\
H9  & Compound-base normalization      & --- (projection)          & decision (fixes live call) \\
H13 & VSC regime                       & VSC frequency priors      & oracle + decision \\
H14 & Undercut window (CRN)            & --- (estimator)           & decision \\
H15 & Pit-stop physics                 & --- (mechanism)           & parity-gated$^{\ast}$ \\
H17 & SC-reactive rival pull-forward   & --- (mechanism)           & parity-gated$^{\ast}$ \\
H18 & Live Kalman pace filter          & live green laps           & parity-gated$^{\ast}$ \\
\bottomrule
\end{tabular}
\caption{Flag-gated engine components and the path each serves after
held-out gating. ``Oracle'' = calibrated outcome probabilities;
``decision'' = recommendations and counterfactuals.
$^{\ast}$Implemented and bit-identical to the base engine at default
settings; enabled on a path only after the pending held-out win
(Section~\ref{sec:dualpath}).}
\label{tab:components}
\end{table}

% -------------------------------------------------------------------------
\section{The Monte Carlo Engine}
\label{sec:mc}

\subsection{Race state and vectorized formulation}
\label{sec:vector}

Let $C$ denote the number of running cars (${\le}20$) and $N$ the number of
Monte Carlo replications ($N{=}2{,}000$ in production). The engine maintains,
for each replication, per-car cumulative race time, current stint index,
tyre age, and pit schedule; all quantities are stored as $N \times C$
arrays. A single simulated lap is one pass of array operations: draw
per-replication track status (green/SC/VSC), compute the $N \times C$
lap-time matrix from the stint arrays (Section~\ref{sec:laptime}), apply
pit losses where the schedule fires (discounted under SC/VSC), apply
dirty-air and overtaking interaction in gap order, and accumulate. Because
strategy plans are precompiled into per-car stint arrays---compound base,
slope, cliff knot, and cliff slope per stint---the inner loop contains no
per-car branching. The vectorized engine is validated against a scalar
reference implementation by distributional parity tests (maximum
per-driver mean-finish discrepancy below $0.6$ positions across the field)
and runs $10$--$50\times$ faster, which is what makes lap-rate re-simulation
of a full field feasible on a single CPU node during a live session.

Retirements are drawn per car per replication with calibrated probability
(\texttt{dnf\_prob}); a retired car is placed behind all classified
finishers. Grid position enters as an initial time spread
(\texttt{grid\_spread} seconds per slot) plus a grid-informed prior on
driver pace, and per-lap Gaussian pace noise (\texttt{lap\_noise}) makes
lap-level outcomes stochastic even under a fixed strategy.

\subsection{Lap time and tyre degradation (H5)}
\label{sec:laptime}

The lap time of a car on compound $t$ at tyre age $a$ is
\begin{equation}
\ell(t, a) \;=\; f_0^{(t)} \;+\; \delta^{(t)} a
\;+\; \ind{a > k^{(t)}}\, s^{(t)} \bigl(a - k^{(t)}\bigr)
\;+\; \epsilon,
\qquad \epsilon \sim \mathcal{N}(0, \sigma^2_{\text{lap}}),
\label{eq:deg}
\end{equation}
where $f_0^{(t)}$ is the fitted compound base pace, $\delta^{(t)}$ the
linear wear slope, and $(k^{(t)}, s^{(t)})$ an optional per-circuit,
per-compound \emph{cliff}: extra degradation of $s^{(t)}$ seconds per lap
beyond the knot age $k^{(t)}$. Two corrections are essential in practice.
First, raw fitted slopes conflate tyre wear with fuel burn-off and track
evolution; because fuel effect and track grip \emph{improve} lap time as the
race progresses, the blended slope is frequently negative---a model in which
tyres improve forever, which a plan optimizer will exploit with impossible
50-lap soft stints. We therefore decompose the slope by adding back a
calibrated per-lap fuel effect and clamping the tyre-only wear component to
a physically plausible band $[0, 0.22]$\,s/lap. Second, the cliff parameters
are \emph{mined, not assumed}: hinge fits on fuel-corrected per-stint deltas
over 8{,}278 stints (191k laps) yield 37 circuit--compound cliffs passing
an evidence gate (${\ge}25\%$ SSE reduction over the linear fit,
$n \ge 300$ laps, physically sane slope). Pooled globally the cliff
signal is survivorship-flattened---stints that hit the cliff are the ones
that ended---so circuit-specific mining is necessary. The decision path
additionally enforces a \emph{max-stint feasibility cap}: no recommended
stint may exceed the longest stint any car demonstrably completed on that
compound at that venue (95th percentile of circuit history pre-race; the
demonstrated maximum post-race), which automatically inherits venue-specific
regulations such as mandated tyre-life limits.

\subsection{Overtaking: the H7 kernel}
\label{sec:h7}

The base engine resolves on-track proximity with a deterministic rule: a car
within the calibrated dirty-air window of the car ahead loses
\texttt{dirty\_air\_loss} seconds per lap and passes only when its pace
advantage exceeds a per-circuit threshold. To test whether a learned
stochastic model improves the simulation, we fitted a logistic overtake
kernel on every adjacent-car battle in the store (a battle: consecutive
cars within the interaction window on a green lap, pit-cycle position
changes excluded): 131{,}000 battles containing 4{,}142 clean passes
(3.2\% base rate). The kernel is
\begin{equation}
P(\pass) \;=\; \sigma\!\bigl(A_{GP} + k_p \,\Delta_{\text{pace}}
 + k_t \,\Delta_{\text{age}}\bigr),
\label{eq:h7}
\end{equation}
with $\Delta_{\text{pace}}$ the pace delta to the car ahead
(s/lap, positive when the attacker is faster), $\Delta_{\text{age}}$ the
tyre-age delta, and $A_{GP}$ an additive-shrinkage per-circuit difficulty
offset. Fitted on seasons ${\le}2023$ (89k battles) and tested on
${\ge}2024$ (41k)---the kernel's own task-level split; the engine-level
gating below uses the engine's held-out seasons
(Table~\ref{tab:ablation})---the kernel is well behaved: standardized coefficients
$k_p{=}{+}0.87$, $k_t{=}{+}0.46$ (both signs correct), monotonic in pace
delta (a car ahead 1.0\,s/lap slower is passed with probability 11\% per
battle-lap; 2.0\,s/lap, 27.5\%), test Brier $0.0275$ against a
climatological baseline of $0.0298$, and passes-per-race MAE $8.9$
(mean 23.5). Circuit offsets recover known difficulty orderings (hardest:
Monaco, Imola, Canada; easiest: Qatar, Spa, Las Vegas).

When wired into the simulator (sampling $\text{rng} \ge P(\pass)$ in
gap order each lap) and fully re-calibrated, however, the kernel does
\emph{not} improve field-level outcome metrics---held-out Brier
$0.0757$ vs.\ $0.0745$ for the threshold engine, with essentially
identical rank correlation (Table~\ref{tab:ablation}). The diagnosis is
that at a $\sim$3\% per-battle base rate the kernel keeps the field almost
as order-locked as the threshold rule, and the calibration optimizer
compensates through the pace-spread parameters in either configuration:
the binding constraint on finishing-order fidelity is the magnitude of
pace separation, not the functional form of the pass model. We report
this as a negative result; the kernel remains available (default-off)
where its per-battle probabilities are the object of interest, e.g.\
live overtake-watch narratives.

\subsection{Compound-base normalization (H9)}
\label{sec:h9}

Per-compound base paces $b_c$ are fitted independently per race, which
confounds a compound's intrinsic pace with \emph{when in the race it was
run}: soft tyres run early on full fuel and a green track, hards late on
low fuel and rubbered-in asphalt. The result is frequent order inversions
(in 18 of 24 races in the 2025 season the soft was \emph{not} the fastest
fitted compound) with spreads up to 3.1\,s/lap---and a live decision
pathology: boxing onto the spuriously ``fast'' compound dominates every
counterfactual, so the engine over-recommends the undercut. A joint
per-race fit $\text{lap} \sim b_c + \delta_c a + \gamma\,\text{racelap}$ is
ill-conditioned for exactly the same reason (compound and race phase are
collinear when a compound spans only one phase), so we use a robust
projection: with $\bar b$ the mean fitted base and $\bar c$ a canonical
centroid respecting soft $\le$ medium $\le$ hard,
\begin{equation}
\hat b_c \;=\; b_c - \bar b + \bar c,
\qquad
\bar c := \arg\min_{c\,\in\,\mathcal{C}} \lVert c - b \rVert^2
\;\;\text{s.t.}\;\;
g_{\min} \le c_{t+1} - c_{t} \le g_{\max},
\label{eq:h9}
\end{equation}
i.e.\ the fitted bases are projected onto the nearest vector with canonical
ordering and adjacent gaps constrained to $[g_{\min}, g_{\max}]$
(calibration bounds $[0, 0.2]$ and $[0.3, 1.2]$\,s/lap respectively), while
the mean pace level and all wear slopes are preserved. On the
decision path this correction is unambiguous: in a representative
pathological case (2025 Spanish GP, lap 46) the recommended box-now call
carried an inflated 72.9\% success probability under raw bases and a
realistic 23.4\%---matching the full probability board---after projection,
flipping the recommendation to stay out. On the oracle path the projection
\emph{hurts}: re-calibrated with normalization enabled, held-out Brier
degrades from $0.0745$ to $0.0784$ (Table~\ref{tab:ablation}), and the
optimizer pushes the projection strength to its floor, indicating that the
confounded raw fit encodes genuine race-specific pace signal
(track-evolution and fuel effects that really did apply to the laps each
compound ran). This asymmetry is the clearest instance of the
calibration/decision split formalized in Section~\ref{sec:dualpath}.

\subsection{Safety Car and Virtual Safety Car (H13)}
\label{sec:h13}

Track-status regimes are simulated per replication per lap. A full Safety
Car deploys with per-lap hazard $p_{\text{SC}} = \lambda_{\text{SC}} /
L_{\text{rem}}$, where $\lambda_{\text{SC}}$ is a per-circuit expected-SC
prior scaled by the calibrated multiplier \texttt{sc\_rate\_mult}; under SC,
lap times are multiplied by \texttt{sc\_pace\_mult} ($\approx 1.4$) and pit
loss by \texttt{sc\_pit\_factor} ($\approx 0.45$)---capturing the cheap-stop
dynamics that dominate real strategy resets. The VSC regime (new in this
system relative to prior academic simulators) is drawn analogously with its
own rate, a shorter duration (uniform on $1..\lfloor
\texttt{vsc\_dur\_max}\rfloor$ laps), a milder pace multiplier
(\texttt{vsc\_pace\_mult}, bounds $[1.10, 1.35]$), and an intermediate pit
discount (\texttt{vsc\_pit\_factor}, bounds $[0.45, 0.85]$); SC takes
priority when both would be active, and the two regimes are mutually
exclusive by construction. Modern F1 sees roughly $0.5$--$0.9$ VSCs per
race, and the distinction matters strategically: a VSC stop saves less than
an SC stop but arrives more often and ends sooner, changing the
option value of staying out.

\subsection{Rival policy (H1)}
\label{sec:h1}

Counterfactual evaluation requires an assumption about what the other 19
cars do. Replaying rivals' \emph{historical} plans is wrong twice over: in
backtests it leaks the future, and in live operation their remaining plan is
unknown. \textsc{Pitwall} assumes rivals play their \emph{pace-optimal
remaining strategy}: for each rival, the plan minimizing deterministic
remaining race time under the current compound model, feasibility caps, and
pit-loss constant. This is a conservative, game-theoretically motivated
choice---the focal car's option value is measured against competent
opposition rather than against the (frequently suboptimal) plans teams
actually ran---and it makes backtest and live evaluation structurally
identical. The known limitation is that rivals do not \emph{react} to the
focal car's move (they do not cover the undercut); Section~\ref{sec:limits}
discusses this sequential-game extension.

\paragraph{SC-reactive pull-forward (H17).} A first, well-identified slice
of rival reactivity is already implemented: real teams pull a scheduled stop
forward when a Safety Car or VSC discounts the pit loss, and an open-loop
rival that ignores a cheap-stop window is systematically pessimistic about
neutralisation-period strategy (in one live case the engine gave zero credit
to an SC-window stop that every competent rival would have taken). Under
H17, a rival whose next scheduled stop lies within \texttt{sc\_pull\_window}
laps and whose tyres are at least \texttt{sc\_min\_age} laps old boxes
immediately when a neutralisation is active; cumulative-lap scheduling
guarantees that pulling a stop forward never shortens later stints, and the
focal car is excluded so its committed plan remains the object of the
counterfactual. The mechanism is reactive to the shared \emph{track state},
not to the focal car's private decision, so it composes with common random
numbers without breaking the CRN coupling of Section~\ref{sec:h14}. Like
H15 it ships parity-gated (bit-exact at its default-off setting, verified
by golden-hash regression) pending a held-out calibration win; the full
covering game---a rival responding to the focal undercut itself---remains
future work (Section~\ref{sec:limits}).

\subsection{The undercut window with common random numbers (H14)}
\label{sec:h14}

The core live decision is not ``pit or not'' but \emph{when}: box now, or
in one or two laps, or stay to the pace-optimal stop. \textsc{Pitwall}
evaluates the option set
$\mathcal{H} = \{h{=}0, h{=}1, h{=}2, \text{stay}\}$, where option $h$ runs
$h$ further laps on the current compound and then boxes onto the cheapest
fresh continuation (itself chosen by deterministic plan search under the
feasibility caps), and ``stay'' is the focal car's H1-optimal remaining
plan. Each option is simulated with an \emph{identical} random seed---the
same Safety-Car draws, the same retirement draws, the same lap-noise
sequence---so that for any pair of options the difference in outcomes is
attributable to the strategy alone. Formally, with $F_h^{(i)}$ and
$R^{(i)}$ the finish positions of the focal car under option $h$ and of
the designated rival in replication $i$ of the common world $\omega^{(i)}$,
the engine reports
\begin{equation}
\widehat{P}_h(\text{ahead}) = \frac{1}{N} \sum_{i=1}^{N}
\ind{F_h^{(i)} < R^{(i)}},
\qquad
\widehat{\Delta}_h = \widehat{P}_h(\text{ahead}) -
\widehat{P}_{\text{stay}}(\text{ahead}),
\label{eq:h14}
\end{equation}
along with expected finish position, and win and podium probabilities per
option. Under common random numbers the variance of $\widehat{\Delta}_h$
contains a $-2\,\mathrm{Cov}(F_h, F_{\text{stay}})$ term that is strongly
negative for positively coupled options \citep{glasserman1992some}; in
practice option deltas at $N{=}2{,}000$ are stable to well under a
percentage point, whereas independently seeded options would require an
order of magnitude more replications for the same contrast precision. The
rival defaults to the nearest running car directly ahead on track---the car
an undercut is \emph{for}.

\subsection{Pit-stop physics (H15)}
\label{sec:h15}

The base engine applies a deterministic pit loss, while the analytical
undercut module charges a cold-tyre out-lap penalty the Monte Carlo engine
did not---an internal inconsistency that biased the simulator optimistic
toward pitting and inflated the H14 stay-vs-box deltas on marginal calls.
H15 closes it with three physically grounded terms: stationary time becomes
$\text{pit\_loss}\cdot f_{\text{neut}} + \mathrm{Exp}(\sigma_{\text{pit}})$,
whose right-tailed exponential noise captures the slow-stop tail that a
symmetric Gaussian would understate; an in-lap overhead
$\delta_{\text{in}}$ is charged on the pitting lap; and an out-lap warm-up
penalty $\delta_{\text{out}}$ is charged on the first lap after every stop
(detected as tyre age zero on a non-initial stint, which correctly excludes
the race start). The deterministic optimizer's stint-time model accepts the
same $\delta_{\text{out}}$, so the plan search and the simulator price a
stop identically. All three parameters default to zero---the engine is
bit-identical to its pre-H15 behaviour until calibration frees them---which
is the parity-gating discipline of Section~\ref{sec:dualpath} applied at
the implementation level.

\subsection{Uncertainty quantification}
\label{sec:bootstrap}

Every reported probability carries a 90\% percentile-bootstrap confidence
interval: the $N$ simulated finish matrices are resampled $B{=}500$ times
along the replication axis, and for each driver and event
$E \in \{\text{win}, \text{podium}, \text{points}\}$,
\begin{equation}
\widehat{W}_d = \frac{1}{N} \sum_{i=1}^{N} \ind{\mathrm{rank}(d, i) \in E},
\qquad
\mathrm{CI}_{90}(\widehat{W}_d) =
\bigl[\,q_{5\%}(\widehat{W}_d^{\ast}),\; q_{95\%}(\widehat{W}_d^{\ast})\,\bigr],
\label{eq:boot}
\end{equation}
where $\widehat{W}_d^{\ast}$ are the bootstrap replicates. Because the
bootstrap reuses the already-simulated matrices, intervals attach to the
live probability board at zero additional simulation cost. These intervals
quantify \emph{Monte Carlo sampling} error, not model error; model error is
addressed by the calibration protocol of Section~\ref{sec:calibration}.

% -------------------------------------------------------------------------
\section{Calibration}
\label{sec:calibration}

\subsection{Parameters and bounds}

The engine exposes a twelve-dimensional core parameter vector
$\theta$---per-lap pace noise, retirement hazard, SC pace and pit-loss
multipliers, the dirty-air window and loss, the pace clip, scale, and
grid spread, the fuel effect, and the SC and overtake rate multipliers
(Table~\ref{tab:bounds})---plus the shaping constants of the gated
components (VSC pace and pit multipliers, compound-projection gaps,
pace-prior weights). Every
parameter is searched inside physically interpretable bounds
(Table~\ref{tab:bounds}); the discrete component switches (H5, H7, H9, H13)
are \emph{pinned} per run and each configuration is calibrated separately,
so a component is never half-adopted by a fractional flag. Bounding is not
cosmetic: bounds keep the fitted engine auditable (an SC pace multiplier of
$1.42$ means something to a strategist; an unconstrained value of $3.7$
would mean the optimizer is laundering a misspecification through the SC
model), and \emph{which bounds the optimizer saturates is itself
diagnostic}---the persistent pinning of \texttt{pace\_clip},
\texttt{pace\_scale}, and \texttt{grid\_spread} at their upper bounds
across configurations is what identified pace-spread magnitude, not the
overtake model, as the binding constraint on order fidelity
(Section~\ref{sec:h7}).

\begin{table}[t]
\centering\small
\begin{tabular}{@{}llll@{}}
\toprule
Parameter & Meaning & Bounds & Default \\
\midrule
\texttt{lap\_noise}        & per-lap pace noise (s)              & $[0.05, 0.40]$ & 0.18 \\
\texttt{dnf\_prob}         & per-car retirement probability      & $[0.00, 0.15]$ & 0.07 \\
\texttt{sc\_pace\_mult}    & lap time under SC vs.\ green        & $[1.15, 1.70]$ & 1.40 \\
\texttt{sc\_pit\_factor}   & pit loss under SC vs.\ green        & $[0.20, 0.80]$ & 0.45 \\
\texttt{dirty\_air\_window}& interaction gap window (s)          & $[0.4, 2.5]$   & 1.2 \\
\texttt{dirty\_air\_loss}  & held-up loss (s/lap)                & $[0.0, 0.8]$   & 0.35 \\
\texttt{pace\_clip}        & cap on driver pace offset (s/lap)   & $[0.3, 2.5]$   & 0.6 \\
\texttt{pace\_scale}       & compression of raw pace deltas      & $[0.2, 1.0]$   & 0.55 \\
\texttt{grid\_spread}      & start spread per grid slot (s)      & $[0.0, 1.5]$   & 0.28 \\
\texttt{fuel\_effect}      & fuel burn-off folded out of deg (s/lap) & $[0.00, 0.15]$ & 0.05 \\
\texttt{sc\_rate\_mult}    & multiplier on per-circuit SC prior  & $[0.3, 3.0]$   & 1.0 \\
\texttt{ot\_rate\_mult}    & multiplier on overtake difficulty   & $[0.3, 3.0]$   & 1.0 \\
\texttt{vsc\_pace\_mult}   & lap time under VSC vs.\ green       & $[1.10, 1.35]$ & 1.22 \\
\texttt{vsc\_pit\_factor}  & pit loss under VSC vs.\ green       & $[0.45, 0.85]$ & 0.65 \\
\texttt{vsc\_dur\_max}     & max VSC duration (laps)             & $[2.0, 4.0]$   & 3.0 \\
\bottomrule
\end{tabular}
\caption{Core calibrated parameters with physically plausible search
bounds. Discrete component switches (\texttt{pace\_model},
\texttt{overtake\_model}, \texttt{compound\_norm}, \texttt{tyre\_cliff},
\texttt{vsc\_rate}) are pinned per configuration and searched separately.}
\label{tab:bounds}
\end{table}

\subsection{Objective and protocol}

For each historical race the engine simulates the field from the grid
(the pre-race oracle) and produces per-driver event probabilities
$p_i$ for win, podium, and points, compared to realized outcomes
$y_i \in \{0,1\}$ over classified drivers. Forecast quality is the Brier
score \citep{brier1950verification}
\begin{equation}
\mathrm{BS} = \frac{1}{n}\sum_{i=1}^{n} (p_i - y_i)^2
\;=\; \underbrace{\mathrm{REL}}_{\text{reliability}}
 - \underbrace{\mathrm{RES}}_{\text{resolution}}
 + \underbrace{\mathrm{UNC}}_{\text{uncertainty}},
\label{eq:brier}
\end{equation}
with the \citet{murphy1973new} decomposition computed over binned forecasts;
reliability diagrams plot empirical frequency against forecast probability
per bin, and ECE summarizes the bin-weighted reliability gap. Because a
strategy engine must get \emph{ordering} right as well as probabilities, the
calibration objective blends rank fidelity with calibration:
\begin{equation}
J(\theta) = \rho_{\text{Spearman}}(\theta)
 - w_{B}\,\mathrm{BS}(\theta) - w_{E}\,\mathrm{ECE}(\theta),
\qquad w_B = 2,\; w_E = 1,
\label{eq:objective}
\end{equation}
where $\rho_{\text{Spearman}}$ is the mean rank correlation between
expected and realized finishing order. $J$ is maximized by
bound-constrained differential evolution over the training split
(seasons 2018--2024, 126 races), with $n_{\text{sims}}{=}400$ per race
per evaluation and multiple seeds to control optimizer noise; the test
split (the 2025--2026 seasons, 29 races at the time of the gating runs) is
never seen by the optimizer and is touched only for final configuration
comparison. Each gated configuration (base, +H7, +H9,
+H5) receives its \emph{own} full calibration run---components are never
evaluated with parameters tuned for a different configuration, which would
bias the ablation against the challenger.

\paragraph{Calibration infrastructure (H16).} Two properties of the search
itself proved as important as the objective. First, \emph{identifiability
logging}: alongside differential evolution the harness supports CMA-ES
\citep{hansen2001cmaes}, whose evolved per-parameter standard deviations
report which parameters the data actually pins down and which remain
prior-dominated---a direct check that a fitted value like
\texttt{sc\_pace\_mult}${=}1.4$ is identified rather than inherited from its
initialization. Second, \emph{season-blocked cross-validation}: a $K$-fold
protocol that trains on $K{-}1$ seasons and evaluates on the held-out season
separates calibration variance from genuine era drift, guarding against the
single-split lottery. Both are implemented and parity-tested; the joint
recalibration that frees the pinned H13/H15/H17 parameters under this
protocol is the designated next experiment, and per the gating rule of
Section~\ref{sec:dualpath} none of those components ships enabled until it
produces a held-out win.

\subsection{The dual-path principle}
\label{sec:dualpath}

The gating experiments (Table~\ref{tab:ablation}) produce a consistent
pattern: components that add realism or fix decision pathologies do not
necessarily improve---and can degrade---outcome calibration, because the
calibration objective rewards whatever correlates with finishing order,
including confounds (H9) and compression artifacts (H7). We therefore adopt
a dual-path rule: \emph{the oracle path runs the configuration that wins
the held-out calibration comparison; the decision path runs the
configuration that wins the held-out decision-fidelity comparison; no
component is enabled on a path without a held-out win on that path's
metric.} Concretely, production probabilities come from the base-engine
configuration (H13 enabled), while recommendations, the strategy-cost
report, and live calls use the feasibility-capped H5 tyre model and the H9
projection. This costs one extra deterministic plan search per decision
query and buys an engine whose probabilities and whose advice are each
separately auditable. We consider making this split explicit---rather than
quietly shipping the prettier model everywhere---a core lesson of the
system. The rule has an implementation-level counterpart, \emph{parity
gating}: every new component ships behind a flag whose default reproduces
the incumbent engine bit-for-bit (verified by golden-hash regression
tests), so adoption is always a measured decision against a held-out
metric rather than a silent behavioral drift. H15, H17, and H18 currently
sit in this state, implemented but inert pending their gating experiments.

% -------------------------------------------------------------------------
\section{The Language Layer: Verifier-Gated Briefings}
\label{sec:language}

Everything the engine computes reaches its audience as prose. The
language layer turns the canonical state and the probability board into
short strategy briefings---``Verstappen leads on 18-lap hards; Norris,
1.4\,s back, has the undercut if he boxes within two laps''---in English,
Spanish, and Portuguese, at every lap, for live and replayed sessions.
The design constraint is asymmetric: an omitted fact costs little, while a
fabricated one, attached to a real athlete's name and published in real
time, is a reputational defect of the same severity class as the
pathological pit calls of Section~\ref{sec:h9}. We therefore treat
faithfulness not as a property to be measured after the fact but as a
gate applied at three points: on every published sentence, on every
fine-tuning target, and on the choice of generator itself.

\subsection{Typed claims and reference-free verification}
\label{sec:claims}

A briefing is decomposed into typed factual claims drawn from a schema of
ten types: \textsc{leader}, \textsc{position}, \textsc{gap-to-leader},
\textsc{battle}, \textsc{pace}, \textsc{tyre}, \textsc{safety-car},
\textsc{lap-number}, \textsc{overtake}, and \textsc{position-change}.
Extraction is available in two regimes: a deterministic multilingual
pattern extractor (surnames resolved to canonical three-letter driver
codes), and an LLM extractor for free-form text. Each claim is verified
against the live state snapshot---and, for the two lap-over-lap types
(\textsc{overtake}, \textsc{position-change}), against the previous
snapshot---yielding \textsc{supported}, \textsc{contradicted}, or
\textsc{unverifiable}; a text's faithfulness is the supported fraction of
its claims and its hallucination rate the contradicted fraction. Because
the grounding source is the state itself, the metric is reference-free:
it requires no gold commentary, which for a live session does not exist.

Coverage determines whether these scores mean anything. The pattern
extractor, written to the template's phrasings, recovers only 0--2 claims
from free-form model output that in fact asserts 4--15 verifiable facts,
collapsing scores to degenerate values (0.00, or a vacuous 1.00 at
$n{=}1$); the LLM extractor recovers 8--15 claims per text and yields
discriminative scores (0.78--1.00 across our development texts). This is
the live counterpart of the coverage-aware argument of
\citet{santillana2026precision}: a verifier that cannot see a claim
cannot falsify it, so extractor coverage is a validity precondition, not
an implementation detail. The verification layer is exercised by a suite
of 36 unit tests spanning all claim types and all three languages; one
audit finding is illustrative of why tests must cover the domain's full
history---the compound vocabulary initially omitted the hypersoft,
ultrasoft, and supersoft compounds used in 2018--2021 ($\sim$150
occurrences in the store), silently converting true tyre claims into
false \textsc{contradicted} verdicts.

\subsection{Faithfulness-gated fine-tuning}
\label{sec:sftgate}

A deterministic template generator over the state is faithful by
construction but rigid. To obtain a generator that is natural without
license to invent, we gate the fine-tuning corpus with the verifier
itself. For 145 training races (2018--2024; the 2025--2026 seasons are
held out here exactly as in the engine calibration), at seven lap
fractions per race and in three languages, a frontier model
(GPT-4o) is prompted with the same structured context the production
generator sees and asked for natural commentary. A candidate is admitted
as a fine-tuning target only if it makes at least one checkable claim
and \emph{zero} contradicted claims under
$\textsc{score}(\cdot)$ with the LLM extractor; a cheap
stopword-based language check additionally rejects responses in the wrong
language (caught in $\sim$5\% of Spanish/Portuguese requests in the
pilot). Failures are retried twice, then fall back to the template---so
every one of the 3{,}045 resulting targets is either rich-and-verified or
faithful-by-construction, and the fine-tuned model has, by design, never
been shown an ungrounded target. Of the corpus, 2{,}494 targets (81.9\%)
are admitted model generations and 551 (18.1\%) are template fallbacks.
This inverts the usual relationship between generation and evaluation:
the faithfulness metric is not applied to the model after training but
determines what the model is permitted to learn from. Section~\ref{sec:nlgeval}
evaluates what this buys---and what it does not.

At serving time the same discipline applies downstream: briefing cards
published to the live dashboard and to social channels carry only claims
that pass verification against the current state, so the deployment-time
guarantee does not depend on the generator being perfect, only on the
verifier's coverage of what it says.

% -------------------------------------------------------------------------
\section{Experiments}
\label{sec:experiments}

\subsection{Backtest: outcome prediction over seven scored seasons}

\begin{figure}[t]
\centering
\includegraphics[width=\linewidth]{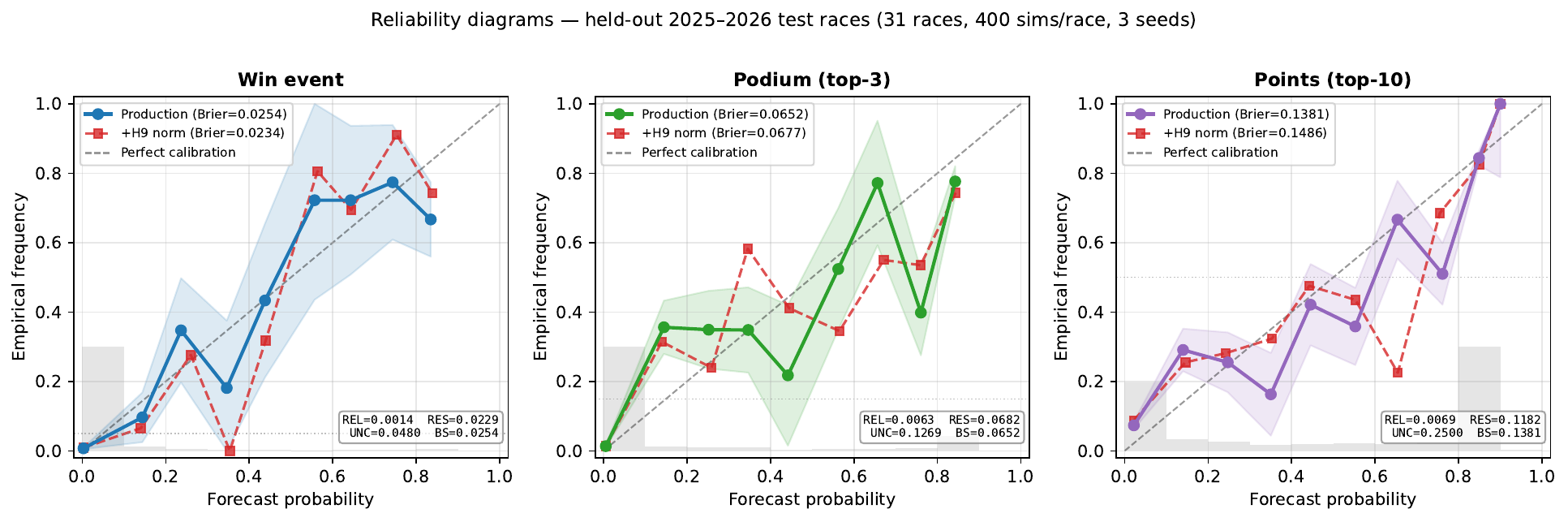}
\caption{Reliability diagrams on the held-out 2025--2026 test races
(31 races, $N{=}400$, 3 seeds; win / podium / points).
Forecast probability on the $x$-axis, empirical event frequency on the $y$-axis;
diagonal = perfect calibration; bar underlay shows bin count.
Production configuration (solid): win Brier $0.0254$ with reliability term
$0.0014$; forecasts sit within a few percentage points of the diagonal in the
well-populated bins, with mild over-confidence in the sparse $0.8$--$0.9$ win
bin. Overlay (+H9 flag-flip on the production parameters, dashed): marginally
better on the win event ($0.0234$) but worse on podium and points (combined
$0.0799$ vs.\ $0.0762$), directionally consistent with the fully recalibrated
gating comparison of Table~\ref{tab:ablation}. Murphy decomposition:
$\mathrm{BS} = \mathrm{REL} - \mathrm{RES} + \mathrm{UNC}$.}
\label{fig:reliability}
\end{figure}

Table~\ref{tab:backtest} reports the pre-race oracle over the full
structured store (155 scoreable races at the time of the backtest;
$n_{\text{sims}}{=}400$ per race), with the calibrated parameter vector. The engine
identifies the eventual winner as its modal winner in 64.5\% of races,
places the winner in its top-3 in 90.3\%, and recalls 72.3\% of podium
finishers, with mean Spearman $\rho = 0.749$ against the final
classification. These full-store figures span training and test years and
are reported for completeness; the held-out 2025--2026 split alone achieves
$\rho = 0.766$ (Table~\ref{tab:ablation}), so the aggregate is not inflated
by in-sample races. Year-over-year stability is good (top-3 between 80.0\% and
100\%; $\rho$ between 0.647 and 0.806) with no evident drift across the
2022 regulation change, indicating that per-race pace fitting absorbs most
era effects. As a reference point, a grid-order baseline (winner = pole)
achieves substantially lower podium recall and rank correlation in every
season; a climatological baseline (championship-order priors) cannot react
to race-specific pace at all. Held-out probability quality for the production
configuration is combined Brier $0.0745$ and ECE $0.0297$ (combined
win / podium / points events, computed in the gating runs on the 29-race
2025--2026 test split, $3$ seeds); re-evaluated on the current 31-race
held-out store (\texttt{calibration\_vec}, $N{=}400$, 3 seeds), per-event
Brier is $0.0254$ for win (reliability term $0.0014$), $0.0652$ for podium,
and $0.1381$ for points.
Reliability diagrams (Figure~\ref{fig:reliability}) show forecasts within a
few points of empirical frequency across the well-populated bins, with mild
over-confidence in the sparsely populated $0.8$--$0.9$ win bin.

\begin{table}[t]
\centering\small
\begin{tabular}{@{}lrrrrr@{}}
\toprule
Season & Races & Winner hit \% & Winner top-3 \% & Podium recall \% & Spearman $\rho$ \\
\midrule
2018 & 20 & 45.0 & 80.0  & 73.3 & 0.705 \\
2019 & 21 & 61.9 & 95.2  & 79.4 & 0.803 \\
2020 & 17 & 76.5 & 82.4  & 70.6 & 0.647 \\
2021 & 22 & 72.7 & 86.4  & 71.2 & 0.747 \\
2023 & 22 & 72.7 & 95.5  & 71.2 & 0.736 \\
2024 & 24 & 62.5 & 95.8  & 73.6 & 0.805 \\
2025 & 24 & 62.5 & 91.7  & 68.1 & 0.757 \\
2026$^\dagger$ & 5 & 60.0 & 100.0 & 66.7 & 0.806 \\
\midrule
Overall & 155 & 64.5 & 90.3 & 72.3 & 0.749 \\
\bottomrule
\end{tabular}
\caption{Pre-race oracle backtest with the calibrated engine
($n_{\text{sims}}{=}400$/race). ``Winner hit'' = modal simulated winner
won; ``winner top-3'' = actual winner within the three most probable.
$^\dagger$Partial season at time of writing.}
\label{tab:backtest}
\end{table}

\subsection{Component ablations on held-out seasons}

Table~\ref{tab:ablation} is the gating experiment: each configuration
independently calibrated on the 2018--2024 training split, evaluated on the
held-out 2025--2026 test split ($n{=}400$ sims $\times$ 3 seeds). The production configuration wins
or ties every calibration metric. Enabling the H7 kernel is calibration-
neutral-to-negative despite the kernel's own out-of-sample validity
(Section~\ref{sec:h7}); enabling H9 normalization or the H5 engine-path
cliff degrades Brier by 4--6\%. We emphasize that these are
\emph{post-recalibration} comparisons---each challenger was given its best
shot---which is what licenses routing those components to the decision path
rather than discarding them.

\begin{table}[t]
\centering\small
\begin{tabular}{@{}lrrrr@{}}
\toprule
Configuration & Brier $\downarrow$ & ECE $\downarrow$ & Spearman $\rho$ $\uparrow$ & Winner top-3 \% $\uparrow$ \\
\midrule
Production (base + H13)            & \textbf{0.0745} & \textbf{0.0297} & \textbf{0.766} & \textbf{95.4} \\
\ + H7 overtake kernel (recal.)    & 0.0757 & 0.0344 & 0.762 & 95.4 \\
\ + H9 compound norm.\ (recal.)    & 0.0784 & 0.0308 & 0.742 & 90.8 \\
\ + H5 cliff, engine path          & 0.0786 & --     & 0.755 & --   \\
\midrule
H7 kernel alone (pass-level task)  & 0.0275 & \multicolumn{3}{l}{vs.\ 0.0298 climatological; passes/race MAE 8.9} \\
\bottomrule
\end{tabular}
\caption{Held-out gating (2025--2026 test split; combined win/podium/points Brier/ECE;
$3$ seeds). Realism components that win their own task (bottom row) do not
transfer to field-level calibration --- the empirical basis for the
dual-path design of Section~\ref{sec:dualpath}.}
\label{tab:ablation}
\end{table}

\subsection{Decision fidelity: strategy recommendations against history}

The decision path is evaluated on 42 dry races from 2024--2025, with
the H5 artifact mined on ${\le}2023$ only; this evaluation exercises the
deterministic plan optimizer rather than the calibrated simulator, so the
relevant holdout basis is the artifact's mining cutoff, and all 42 races are
out-of-sample with respect to the mined caps and cliffs.
Table~\ref{tab:decision} compares the recommended plan to the executed race. Without feasibility
constraints the optimizer recommends a stint longer than any real car
achieved in 90.5\% of races---the negative-slope pathology of
Section~\ref{sec:laptime}; with the mined caps and cliff, infeasible
recommendations drop to \textbf{zero}, stop-count agreement with the
modal executed strategy rises from 59.5\% to 69.0\%, and stop-count MAE
falls from 0.40 to 0.31. The baseline's 59.5\% is itself degenerate---it
recommends one stop nearly always; restricted to the 18 races where the
modal real strategy was two stops, the baseline matches 1/18 and the H5
decision path 9/18. Strategy-cost audits corroborate the fix: the median
gap between executed and recommended plan time on pathological venues
collapses (Monza $+73$--$102$\,s $\rightarrow$ median $16$\,s; Suzuka
$\rightarrow 7$\,s; Monaco $\rightarrow 19$\,s; field median of race
medians $11$\,s), and residual high-cost venues are precisely the
track-position circuits (e.g.\ Singapore) where lap-time optimality is the
wrong objective---the report flags these with a confidence gate rather
than feigning precision.

\begin{table}[t]
\centering\small
\begin{tabular}{@{}lrrr@{}}
\toprule
Decision metric (42 held-out dry races) & Base optimizer & +H5 (train-mined) & +H5 (full) \\
\midrule
Recommends an infeasible stint (\%) $\downarrow$ & 90.5 & 50.0 & \textbf{0.0} \\
Stop-count match with executed modal plan (\%) $\uparrow$ & 59.5$^{\ast}$ & \textbf{69.0} & --- \\
Stop-count MAE $\downarrow$ & 0.40 & \textbf{0.31} & --- \\
Match on modal-2-stop races $\uparrow$ & 1/18 & --- & \textbf{9/18} \\
\bottomrule
\end{tabular}
\caption{Decision-path gating. $^{\ast}$Degenerate: the unconstrained
baseline almost always recommends one stop. ``Train-mined'' caps use
${\le}2023$ data only (honest held-out); ``full'' uses all mined circuit
history as deployed.}
\label{tab:decision}
\end{table}

\subsection{Undercut window behaviour and empirical grounding (H14)}

The H14 window's value is contrast precision. With common random numbers
the sign of $\widehat{\Delta}_h$ (Eq.~\ref{eq:h14}) is stable across seeds
at $N{=}2{,}000$ for gaps outside $\pm 0.3$ percentage points, and the
window exhibits the qualitatively correct phenomenology on historical
replays: box-now dominates when the rival ahead is on old rubber near a
mined cliff knot and pit loss is about to be VSC-discounted; stay-out
dominates under high SC hazard (banking a possible cheap stop) and on
high-difficulty circuits where track position outweighs tyre delta. On
undercut-heavy historical moments the H9-projected window agrees with the
full probability board, where the raw-base window systematically inflated
box-now by tens of points (Section~\ref{sec:h9}).

To ground the H14 predictor empirically, we extract all $2{,}134$ undercut
events from \texttt{pit\_battles} across the structured store
($1{,}306$ training events ${\le}2023$; $828$ test events from 2024--2025,
held out with respect to this event model's own fitting cutoff) and
score a marginal model for position-swap success.  We fit a logistic
regression on three inputs available before the stop: (i)~\texttt{gap\_before\_s}
(gap to the target rival in seconds), (ii)~tyre-age delta (attacker age
minus defender age, derived from stint records), and (iii)~log remaining laps.
On the held-out split the model achieves Brier $0.168$ against a
climatological baseline of $0.188$ (Brier Skill Score $0.107$;
position-swap base rate $25\%$, $N{=}828$).  The dominant coefficient is
\texttt{gap\_before\_s} ($\hat{w}{=}{-}1.21$, negative as expected:
a larger gap before the stop makes an outright position swap less likely)
and tyre-age delta is positive but weak ($\hat{w}{=}{+}0.01$), consistent
with the H7 finding that the overtake model captures relatively little
marginal information beyond pace separation.  These results confirm that undercut outcomes are statistically predictable
from pit-time features available without telemetry (BSS~$0.107$), and
independently establish \texttt{gap\_before\_s} as the dominant input.
Note that this logistic evaluation validates \emph{feature importance};
the inter-option deltas $\widehat{\Delta}_h$ from the MC engine are
separately variance-controlled by the shared CRN seed, ensuring they
reflect strategy choice rather than simulation noise.
Both findings motivate H9's role in purging spurious gap estimates from
the live compound-base fit.

\subsection{Live deployment: the 2026 Austrian Grand Prix}
\label{sec:live}

\textsc{Pitwall} operated end-to-end during the 2026 Austrian Grand Prix:
the SignalR live-timing stream was ingested at 1\,s cadence through a
proxied edge agent, decoded into the canonical state (compound, tyre age,
gaps, track status---including correct discrimination of yellow flags from
VSC deployment, validated earlier on a real yellow$\rightarrow$VSC$
\rightarrow$clear sequence in Barcelona free practice), and the full board
(win/podium/points with CIs), the undercut window, and grounded commentary
cards were recomputed continuously as laps arrived. Because 2026 races have
no same-season historical store, compound models were seeded from the 2025
running of the same event and refreshed from live green laps as the race
developed---the pre-race prior mattered mainly for the first stint. The
deployment also functioned as an adversarial test of the state layer: a
mid-race anomaly (a P8 car $+25$\,s showing an implausible 55.8\% win
probability at lap 17/71) was traced to out-lap contamination of the live
pace fit and a compound double-count in the state builder, both fixed and
regression-tested; we report this because live operation surfaces
state-estimation failure modes that no backtest exercises, and the
credibility of a live decision system rests on treating them as
first-class defects rather than presentation glitches. The
end-to-end lap-to-updated-board latency was dominated by the feed cadence
rather than by simulation ($N{=}2{,}000$ full-field continuations complete
in seconds on a single CPU container), confirming that calibrated
counterfactual advice at live cadence is feasible without team-grade
compute.

\paragraph{From incident to estimator (H18).} The out-lap anomaly also
exposed a structural weakness: the live pace offset was a single-lap
estimate ($\sigma \approx 0.18$\,s), so one contaminated observation could
move the entire probability board. The systematic fix is a per-car
two-state Kalman filter over $[\beta_0, \beta_1]$---pace offset (s/lap) and
residual degradation slope (s/lap$^2$)---with observation model
$H = [1, a]$ at tyre age $a$, measurement noise inflated $4\times$ when the
car runs in dirty air, out-laps and neutralised laps excluded from updates,
and a compound change resetting the slope's variance to its prior while
carrying the offset. The filter's posteriors enter the simulator not as
point estimates but as per-replication draws (sampled once outside the lap
loop), so live \emph{estimation} uncertainty propagates into the reported
probabilities alongside Monte Carlo sampling uncertainty. Consistent with
the gating discipline, the filter ships parity-gated---bit-identical to the
single-lap estimator until enabled---and its live validation at a
subsequent Grand Prix is the designated acceptance test.

% -------------------------------------------------------------------------
\subsection{Live deployment II: the 2026 British Grand Prix}
\label{sec:live2}

A second race-distance deployment, one week later at the 2026 British
Grand Prix (Silverstone), exercised the same stack under a different
failure surface and added one capability: authenticating the SignalR
connection against a broadcast subscription unlocked the
positional-telemetry topics that the anonymous connection does not
deliver, enabling a live track map alongside the probability board.

\paragraph{From incident to guardrail: prior contamination.}
The pre-race compound prior follows the recipe of
Section~\ref{sec:live}: with no same-season store for the circuit, base
pace and degradation per compound are seeded from a prior running of
the same event. The 2025 British Grand Prix---the natural analog---was
a wet race, and the fit silently inherited its regime: HARD fitted at
$108.3$\,s/lap against MEDIUM at $94.1$\,s/lap, a $14$\,s/lap spread
where dry slick compounds are separated by $1$--$3$\,s/lap. The live
symptom was unmistakable once it fired: a P21 car, $75.8$\,s off the
lead on ten-lap-old hards, reported an $89.1\%$ win probability---but
only under the candidate plans that pitted it onto the faster-fitted
compound, while every other plan for the same car correctly read
${\sim}0\%$. In the simulator's arithmetic, switching to a compound
$14$\,s/lap faster is time travel, and the plan search found it.
Notably, enabling H9 live did \emph{not} repair the board:
mean-centering re-scales the corrupted bases but carries no information
about which fit is wrong, and the projected prior was wrong by a
comparable margin in the opposite direction (every candidate plan
converged to ${\sim}90\%$). This is consistent with H9's
characterization in Section~\ref{sec:h9}: a projection for implausibly
\emph{ordered} fits of roughly correct scale, not a repair for a prior
estimated under a different weather regime. The operational fix was
selection, not correction: re-seeding from the most recent \emph{dry}
running of the circuit (2024; fitted slick spread $1.2$\,s/lap, with
2023 at $1.3$\,s/lap confirming the scale) restored the board without
interrupting ingestion---the same car fell to $0.0\%$ win, median P18.
The incident yielded a guardrail in the same spirit as the H5
feasibility audit: the analog prior is now gated on the fitted spread
of slick base times (a spread above ${\sim}3$--$4$\,s/lap rejects the
analog) \emph{before} the race, so a physically impossible prior is
caught by an audit rather than discovered on a live board. Taxonomically
this is a second, distinct failure family---prior selection, where
Austria's was state estimation---and, like the first, it was surfaced
only by live operation.

\paragraph{A live probability trace, scored against the official result.}
A lightweight collector polled the public field board every $120$\,s
and appended timestamped snapshots to disk; it was deployed mid-race,
so the trace covers the closing quarter---14 polls over laps 39--52 of
52 (lap 51 was sampled twice; the identical repeat is shown once in
Table~\ref{tab:silverstone}). We score each poll's full-field win
vector against the official race outcome. The trace splits into three
regimes. Laps 39--41 straddle the front-runners' pit cycle and---as
became public only after the race---the championship leader's
developing car failure (a wheel-shield detachment on lap 41 that forced
two further stops): the top pick alternated among the three
front-runners, and the per-poll field win-Brier ran $0.071$--$0.087$.
This window is a live instance of declared limitation (i) of
Section~\ref{sec:limits}: a mechanical failure is invisible to the
public timing stream until it manifests in lap time, and no simulator
conditioned on that stream could have priced it. From lap 42 the board
locked onto the eventual winner at $97.7$--$99.5\%$, and the per-poll
field win-Brier fell below $10^{-3}$ in every subsequent poll but one:
a single transient at lap 50 ($55.4\%$, win-Brier $0.014$) under the
race-closing Safety Car, when the bunched field compresses exactly the
gap structure that separates win-probability mass; the board recovered
by the following poll and the race finished under that Safety Car.

The classification coda is instructive. At the flag, the board's
on-road ordering matched the official classification exactly through
P8; the single divergence was the failed former front-runner, ninth on
the road but sixteenth in the official classification after a
five-second track-limits penalty was applied to a field bunched behind
the Safety Car. The simulator predicts on-road order; stewarding
decisions that reorder it after the flag are a noise channel the
backtest oracle silently charges against the engine
(limitation~(vii)). Fourteen polls from one race are an anecdote, not
calibration evidence---the reliability claims of this paper rest on
Section~\ref{sec:experiments}---but the trace is the system's first
end-to-end live artifact in which every number was committed to disk
before the outcome was known.

\begin{table}[t]
\centering\small
\begin{tabular}{@{}rcrrr@{}}
\toprule
Lap (of 52) & Caution & Top pick & Win\,\% & Field win-Brier \\
\midrule
39 & \checkmark & ANT & 88.1 & 0.071 \\
40 &            & HAM & 96.2 & 0.087 \\
41 &            & ANT & 92.2 & 0.077 \\
42 &            & \textbf{LEC} & 97.7 & ${<}10^{-3}$ \\
44 &            & \textbf{LEC} & 97.8 & ${<}10^{-3}$ \\
45 &            & \textbf{LEC} & 98.7 & ${<}10^{-3}$ \\
46 &            & \textbf{LEC} & 98.2 & ${<}10^{-3}$ \\
47 &            & \textbf{LEC} & 99.0 & ${<}10^{-3}$ \\
48 & \checkmark & \textbf{LEC} & 99.0 & ${<}10^{-3}$ \\
49 & \checkmark & \textbf{LEC} & 99.0 & ${<}10^{-3}$ \\
50 & \checkmark & \textbf{LEC} & 55.4 & 0.014 \\
51 & \checkmark & \textbf{LEC} & 99.5 & ${<}10^{-3}$ \\
52 & \checkmark & \textbf{LEC} & 100.0 & ${<}10^{-3}$ \\
\bottomrule
\end{tabular}
\caption{Live probability trace, 2026 British Grand Prix: the field
board's top win pick at each $120$\,s poll, laps 39--52. LEC (bold) is
the eventual winner; the board locks onto him at lap 42 and never flips
thereafter. ``Caution'' is the neutralisation flag (SC/VSC) as ingested
in the same snapshot. Field win-Brier scores the full 22-car win vector
of each poll against the official outcome. The final poll coincides
with the chequered flag. Raw snapshots:
\texttt{live\_predictions\_silverstone.jsonl}.}
\label{tab:silverstone}
\end{table}

% -------------------------------------------------------------------------
\subsection{Briefing generation: the richness--precision trade-off}
\label{sec:nlgeval}

Two generator adapters were fine-tuned from the same base model
(Qwen2.5-3B with LoRA) and differ only in their targets: \textbf{v1}
was trained on the deterministic template outputs verbatim, \textbf{v2}
on the faithfulness-gated corpus of Section~\ref{sec:sftgate}
(admitted rich targets plus template fallbacks). We audited both on a
deployment-shaped probe set: real production contexts from a held-out
2025 race at early, middle, and late race stages, plus constructed
\emph{sparse} contexts (a single ranked car, as arises in early laps or
after an ingestion gap) and end-of-race contexts, under greedy decoding
at two token budgets. We report this as a qualitative audit with
deployment consequences rather than a large-scale benchmark; the
companion paper \citep{santillana2026precision} carries the quantitative
evaluation programme.

The outcome is a clean dissociation. \textbf{v1 is faithful but terse}:
on every probe, including the sparse ones, it terminates cleanly at
28--52 tokens, makes no contradicted claims, and adds little beyond the
template. \textbf{v2 is richer and remains grounded when the context is
dense}---with six ranked cars its output is genuinely more vivid than
v1's while every claim verifies---but on sparse contexts it fills its
learned register from the prior: given only the leader and his tyre age,
it invented pursuing drivers, gap figures, and compound states out of
whole cloth (``closing in at 1.8\,s on fresher mediums'' with no such car
in the context), in direct violation of its system prompt. Notably, the
faithfulness gate on the training data did not prevent this: every v2
target was verified against \emph{its own} context, but rich targets
teach a rich register, and at inference the register demands more facts
than a sparse state supplies. Grounded targets are therefore necessary
but not sufficient; the density mismatch between training and inference
contexts is a distinct failure axis.

To test whether this failure is intrinsic to faithfulness-gated
fine-tuning, we repeated the recipe---same gated corpus, same LoRA
configuration---on three further base models spanning an order of
magnitude in scale, Llama-3.2-1B, Phi-4-mini (3.8B), and
Mistral-7B-Instruct, and re-ran the same dense-and-sparse audit. The
sparse-context failure replicates on two of the three: the 1B model
invents a driver who does not exist in Formula~1 at all, and the 7B
model---the one that fit the training corpus best, with the lowest final
training loss---fabricates real but absent drivers most fluently,
suggesting that fitting the rich register harder makes the density
mismatch worse, not better. Phi-4-mini is the exception: on the same
sparse probes it uniquely degrades to generic, non-checkable phrasing
(``the rest of the field chasing hard'') instead of inventing entities,
while remaining as vivid, and as fully verified, as the Qwen v2 adapter
on dense contexts. Faithfulness under sparsity is therefore not monotone
in parameter count (the 1B and 7B bases fail where the 3.8B succeeds)
and is not conferred by the gated corpus alone; it tracks the base
model's instruction adherence. This is the third gate promised at the
top of Section~\ref{sec:language}---the choice of generator itself---and
it ranks with data gating: an ungated corpus cannot be rescued by a
compliant base, and a gated corpus does not rescue a non-compliant one.

The deployment consequence follows the dual-path rule of
Section~\ref{sec:dualpath}: a rich generator is admissible on the live
path only if it survives the sparse audit. The production commentary
model is accordingly the Phi-4-mini adapter, serving under the
downstream claim verification of Section~\ref{sec:sftgate}; richer bases
that fail the audit remain admissible only behind a context-density gate
(a minimum count of ranked cars in the state) or after retraining with
sparse-context abstention targets---mirroring, in the language layer,
exactly the routing logic that keeps the H9 projection off the oracle
path. Sparse faithfulness also proved fragile to serving-time choices,
in ways the audit caught before deployment: sampled decoding at
temperature $0.7$ hallucinates where greedy decoding does not (we serve
greedy); quantizing the merged model to 8~bits preserves the abstention
behaviour, while 4-bit quantization introduces compound errors and
applying the LoRA adapter over an already-quantized base reintroduces
invented drivers. We also note the audit corrected our own
records: an earlier claim that v1 degenerates into repetition on
late-race contexts did not reproduce under either token budget, a
reminder that negative folklore needs re-verification as much as
positive results do.

% -------------------------------------------------------------------------
\section{Discussion}
\label{sec:limits}

\paragraph{Limitations.}
\textsc{Pitwall} sees only what the public timing stream carries.
(i)~\emph{No car telemetry}: retirements are a calibrated hazard rate, not
a mechanical model; brake or power-unit degradation that a team would see
coming is invisible until it manifests in lap time. (ii)~\emph{Fuel mass is
approximate}: the fuel effect is a calibrated per-lap constant folded out of
degradation, not a measured load; systematic fuel-saving phases can bias
short-horizon pace estimates. (iii)~\emph{Tyre thermals are only partially modeled}:
the degradation model is age-based; H15 parametrizes the cold-tyre out-lap
and the slow-stop tail, but thermal degradation versus wear and
compound-specific working windows remain absorbed into noise and the mined
cliff, and the H15 constants await calibration. (iv)~\emph{Rain is out of
scope}: wet--dry crossover strategy requires a switching model and live grip
estimation; the engine currently gates its confidence rather than advising
in mixed conditions.
(v)~\emph{Rivals react only to track state}: H17 lets rivals take cheap-stop
windows under SC/VSC, but the H1 policy remains open-loop with respect to
the focal car; modeling the covering game (a rival responding to the focal
undercut within one lap) is the natural sequential extension, and our
evidence suggests it---not compound pace---is the main residual error source
in undercut-heavy moments. (vi)~\emph{Calibration inherits the store}: eras with sparse or
inconsistent timing data (notably 2022 in our store) contribute less; and
the pre-race oracle conditions on the grid, so it measures strategy-and-race
simulation quality, not qualifying prediction.
(vii)~\emph{The simulator predicts on-road order}: post-race stewarding
can reorder the official classification (at the 2026 British Grand Prix a
five-second penalty applied to an SC-bunched field cost one car seven
places, Section~\ref{sec:live2}); the backtest oracle grades against the
official result, so this noise is charged to the engine. Mining penalty
frequencies into an explicit classification-noise term, alongside
\texttt{dnf\_prob}, is the planned treatment.

\paragraph{Ethical considerations.}
All outputs are probabilistic forecasts with explicit sampling uncertainty,
presented for analysis, education, and broadcast-style commentary; they are
not deterministic predictions, and the system is not designed for, offered
for, or connected to wagering. Raw feed data is redistributed to no one:
the public artifacts are derived structured summaries, model parameters,
and code, consistent with the data provider's terms. The commentary layer
publishes only claims that pass verification against the engine state,
limiting the reputational harm of confidently wrong generated statements
about real, named athletes.

% -------------------------------------------------------------------------
\section{Conclusion}
\label{sec:conclusion}

We presented \textsc{Pitwall}, a production system that generates
faithful, trilingual natural-language race-strategy briefings from a
calibrated real-time Monte Carlo engine: probabilities are calibrated,
recommendations are feasibility-audited, published sentences are verified
claim-by-claim against the state that prompted them, and every model
component---simulation or generation---carries a held-out justification
for the path it serves. The system demonstrates that the core of a
competitive pit-wall decision tool---calibrated win probabilities,
variance-controlled undercut counterfactuals, SC/VSC option pricing, and
verifier-gated verbalisation---can be built, validated against seven
scored seasons of historical data, and confirmed operationally feasible
in live deployment at consecutive Grands Prix, from public data on
commodity infrastructure. The
recurring empirical lesson is
that realism, task-level accuracy, decision quality, and generation
richness are different
objectives that must be measured separately: the components that look most
like domain knowledge (a learned overtake model, canonical compound
ordering) are exactly the ones our gating protocol declined to put on the
calibrated path, and the generator trained on richer verified targets is
exactly the one that hallucinates when the grounding state thins---unless
the base model supplies, through instruction adherence, the abstention
that gated data alone does not. The path to team-grade deployment is clear and mostly a
data problem: closing the loop on rival reactions (the covering game, of
which the SC-reactive pull-forward of H17 is the first, already-implemented
step), replacing the fuel and thermal constants with telemetry-informed
estimates (H15 supplies the parametrization), extending the regime model to
wet crossovers, and hardening the live state estimator (the H18 Kalman
filter) and the pre-race priors (the analog spread gate of
Section~\ref{sec:live2}) against the failure modes that only live
operation reveals. We intend the calibration protocol---bounded parameters, per-path
gating, and published negative results---as a template for decision-support
systems in sports beyond racing.

% -------------------------------------------------------------------------
\section*{Reproducibility}
The structured race store schema, engine code (vectorized and scalar
reference), calibration harness, mined H5/H7 artifacts, and backtest
scripts are available in the project repository; raw timing data is not
redistributed and is re-derivable from public sources via the documented
ETL. All held-out splits, seeds, and bounds used in
Tables~\ref{tab:ablation}--\ref{tab:decision} are specified in the
repository's experiment logs.

% -------------------------------------------------------------------------
\bibliographystyle{plainnat}

\end{document}